\documentclass[10pt,twocolumn,letterpaper]{article}
\usepackage{geometry}
\usepackage{paper}
\usepackage{times}
\usepackage{epsfig}
\usepackage{graphicx}
\usepackage{amsmath}
\usepackage{amssymb}

\usepackage[linesnumbered, ruled]{algorithm2e}
\usepackage{algorithmic}
\usepackage[utf8]{inputenc} 
\usepackage[T1]{fontenc}    
\usepackage{url}            
\usepackage{booktabs}       
\usepackage{amsfonts}       
\usepackage{nicefrac}       
\usepackage{microtype}      
\usepackage{amsmath,amssymb}
\usepackage{color}
\usepackage{float}
\usepackage{array}
\usepackage{multirow}
\usepackage{booktabs}
\usepackage{subfigure}
\usepackage{amsthm}
\usepackage{wrapfig,lipsum,booktabs}
\usepackage{xcolor}
\usepackage{bm}
\usepackage{textcomp}
\usepackage[noblocks]{authblk}
\usepackage[breaklinks=true,bookmarks=false, colorlinks, linkcolor=red]{hyperref}

\cvprfinalcopy 

\def\cvprPaperID{2305} 

\begin{document}
\newgeometry{top=2.2cm,bottom=2.2cm, left= 1.9cm, right= 1.9cm}

\title{Ensembles of Deep Neural Networks for Action Recognition in Still Images}

\author[ ]{\vspace{-1.0cm}Sina~Mohammadi}
\author[ ]{Sina~Ghofrani~Majelan}
\author[ ]{Shahriar~B.~Shokouhi\vspace{-0.3cm}}

\affil[ ]{\small School of Electrical Engineering, \\Iran University of Science and Technology, Tehran, Iran}

\affil[ ]{\tt\small sina.mhm93@gmail.com, sghofrani1@gmail.com, bshokouhi@iust.ac.ir\vspace{-0.65cm}}


\maketitle
\thispagestyle{empty}

\begin{abstract}
Despite the fact that notable improvements have been made recently in the field of feature extraction and classification, human action recognition is still challenging, especially in images, in which, unlike videos, there is no motion. Thus, the methods proposed for recognizing human actions in videos cannot be applied to still images. A big challenge in action recognition in still images is the lack of large enough datasets, which is problematic for training deep Convolutional Neural Networks (CNNs) due to the overfitting issue. In this paper, by taking advantage of pre-trained CNNs, we employ the transfer learning technique to tackle the lack of massive labeled action recognition datasets. Furthermore, since the last layer of the CNN has class-specific information, we apply an attention mechanism on the output feature maps of the CNN to extract more discriminative and powerful features for classification of human actions. Moreover, we use eight different pre-trained CNNs in our framework and investigate their performance on Stanford 40 dataset. Finally, we propose using the Ensemble Learning technique to enhance the overall accuracy of action classification by combining the predictions of multiple models. The best setting of our method is able to achieve 93.17$\%$ accuracy on the Stanford 40 dataset.

\end{abstract}

\section{Introduction}

Human action recognition has been an active area of research in computer vision and pattern recognition in recent years. The recognition of human actions from still images has useful applications, including image annotation~\cite{xu2015show}, image and video analysis~\cite{tran2015learning}, and Human-Computer Interaction (HCI)~\cite{chi2016enhancing}. In contrast to the video-based action recognition methods~\cite{wu2014towards,zhang2016efficient} which utilize motion cues, still image-based action recognition~\cite{guo2014survey} methods use statistic cues. Therefore, the process of recognizing actions from videos is not applicable to still images. Even though numerous action recognition methods have been proposed over the last decade, action recognition in images is still challenging as a consequence of the viewpoint variations, complicated backgrounds, and variations in human pose. 

The traditional approach to action recognition in still images is the Bag of Visual Words (BoVW)~\cite{peng2016bag,ullah2010improving,oneata2013action} which is capable of acquiring a global representation of an image. Delaitre et al.~\cite{delaitre2010recognizing} classified human actions by adopting a BoVW and an SVM classifier. Some methods modeled human-object interactions for recognizing human activities. Yao et al~\cite{yao2010modeling} used pose information and the objects to model human-object interaction. Prest et al.~\cite{prest2011weakly} used a method to learn the relationship between objects and humans, by adopting a weakly supervised learning scenario. Some works used part-based methods which fuse global features with features of different body parts to recognize human actions~\cite{zhao2017single, gkioxari2015actions}.

In recent years, by virtue of the spectacular success of Convolutional Neural Networks (CNNs) in computer vision~\cite{krizhevsky2012imagenet, girshick2014rich, girshick2015fast} and their powerful feature extraction capability from raw images, deep learning has appeared as a promising approach for  recognizing human actions. Gkioxari et al.~\cite{gkioxari2015actions} classified human actions and attributes by integrating CNNs and Poselets. Oquab et al.~\cite{oquab2014learning} adopted the transfer learning technique by just retraining the classifier of a pre-trained network. Yan et al.~\cite{yan2017multibranch} adopted VGG16~\cite{simonyan2014very} network along with two additional attention branches for recognizing actions in still images.

Despite the great success of CNNs, the lack of large amounts of labeled data in action recognition causes problems for training deep CNNs due to the overfitting issue. In this paper, by benefiting from the CNNs pre-trained on ImageNet, we adopt the transfer learning technique to overcome the lack of large labeled action recognition datasets. Furthermore, it is evident that the deepest layer of the CNN has class-specific information due to its large receptive field. In order to extract more powerful features for action classification, we apply an attention mechanism on the output feature maps of the CNN to adaptively weight the channels of these feature maps. Moreover, we adopt eight different pre-trained CNNs in our framework. To further improve the action classification accuracy, we adopt the Ensemble Learning technique in which the decisions from multiple models are combined. Our contributions are three-fold:

\begin{figure*}[!bt]
\begin{center}
\includegraphics[width=1\linewidth]{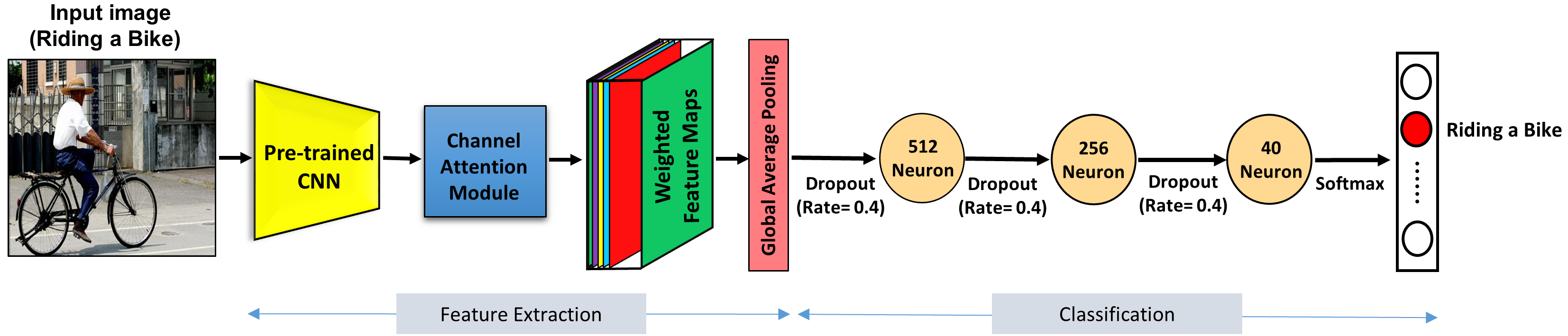}
\end{center}
\caption{The overall pipeline of the proposed method. Our framework consists of two main components: i) The Feature Extraction part which extracts discriminative features from raw images and ii) the Classification part which takes a feature vector as input and generates the probabilities for 40 classes of the Stanford 40 dataset. Note that in the Pre-trained CNN, all fully connected layers are removed.}
\label{fig:main}
\end{figure*}

\begin{itemize}
\item We employ the transfer learning strategy to tackle the lack of labeled data.

\item We propose using an attention mechanism on top of the CNN to extract more discriminative features.

\item We investigate eight different pre-trained CNNs in our framework and propose adopting the Ensemble Learning technique to enhance the overall accuracy of action classification by combining the predictions of different models.  
\end{itemize}

\begin{figure}[!t]
\begin{center}
\includegraphics[width=1\linewidth]{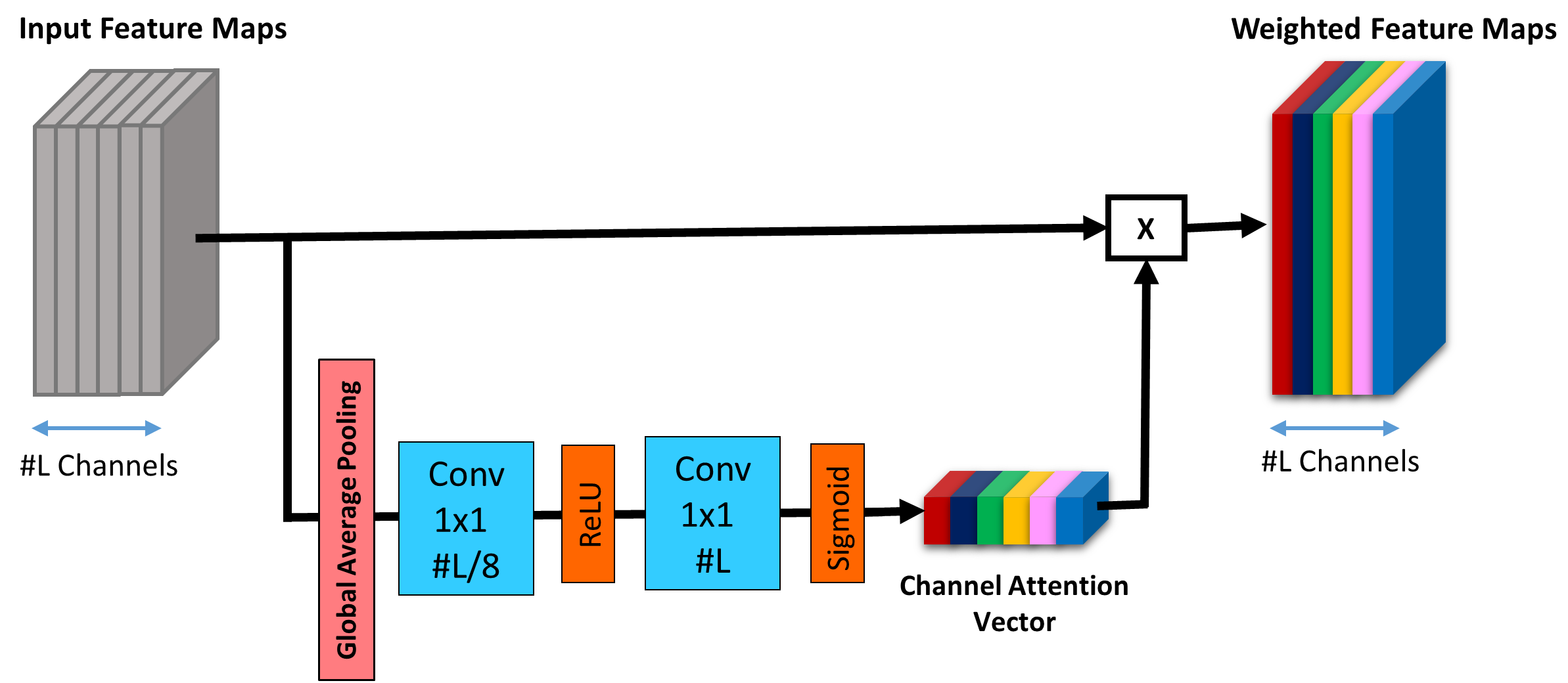}
\end{center}
\caption{The illustration of the Channel Attention Module. This design computes a channel attention vector to re-weight the input feature maps.}
\label{fig:CA}
\end{figure}
	
\section{The Proposed Approach}
We overcome the lack of massive labeled action recognition image dataset by adopting the transfer learning technique. We propose an approach that employs pretrained CNNs designed for solving a different classification task, namely the ImageNet challenge~\cite{ILSVRC15}. Transfer learning takes advantage of knowledge from the source task (ImageNet challenge) to improve learning in the target task (action recognition). In this paper, by exploiting the transfer learning technique, we aim at improving the classification accuracy in Stanford 40 dataset~\cite{yao2011human}. We use eight different CNNs pre-trained on ImageNet, and by removing their fully connected layers and designing some layers on top of them, we build our architecture for action classification. The proposed framework, which is depicted in Fig.~\ref{fig:main}, is composed of two parts, namely the Feature Extraction part and the Classification part.

\subsection{The Feature Extraction Part}
Our proposed Feature Extraction part is composed of a pre-trained CNN, a Channel Attention Module, and a Global Average Pooling layer. As previously described, we adopt the pre-trained CNN to tackle the lack of labeled action recognition datasets. In order to equip our feature extraction part with the power of extracting more discriminative features, we use an attention mechanism on top of the CNN. This attention mechanism is implemented by adopting the Channel Attention Module which is similar to the recently introduced Squeeze and Excitation Networks~\cite{hu2018squeeze}. This module is able to adaptively weight the channels of feature maps resulted from the CNN to select more powerful features in order to use by the Classification part. The reason behind using the Channel Attention Module is that the last layer of the CNN captures global context of the image due to its large effective field of view, and thus the last layer of the CNN has class-specific information. Therefore, by adopting the attention mechanism on top of the CNN, the Feature Extraction part can extract more powerful features for the Action Classification part. The Channel Attention Module is illustrated in Fig.~\ref{fig:CA}.  By adopting some convolutional layers and activation functions, this design computes a channel attention vector to re-weight the input feature maps.

\subsection{The Classification Part}
The Classification part is composed of some fully connected and dropout layers. In the Feature Extraction part, by performing Global Average Pooling, a feature vector is obtained, which is then fed to multiple fully connected and dropout layers. Dropout layers are used to reduce overfitting. We empirically set the dropout rate to 0.4. Finally, the Softmax activation is applied to the output of 40 neurons to obtain the probabilities for 40 classes of the Stanford 40 dataset. For action classification task, we utilize the Cross-entropy loss which is formulated as: 

\begin{equation}
Cross\ Entropy=-\frac{1}{N}\sum_j{y_{j}\times\log_{}{\hat{y_{j}}}}
\end{equation}
\noindent where $y$, $\hat{y}$, and  $N$ denote the ground truth, the prediction, and the number of training examples respectively.

\subsection{Ensemble Learning}
Ensemble Learning is a technique in machine learning which aims at improving the overall performance by combining the decisions from multiple networks. We investigate eight different pre-trained CNNs, namely VGG-16~\cite{simonyan2014very}, ResNet50~\cite{he2016deep}, Inception V3~\cite{szegedy2016rethinking}, InceptionResNetV2~\cite{szegedy2017inception}, DenseNet201~\cite{huang2017densely}, Xception~\cite{chollet2017xception}, NASNet-Mobile~\cite{zoph2018learning}, and NASNet-Large~\cite{zoph2018learning}. We train our proposed model with each of these eight CNNs, and then we apply the Ensemble Learning technique to the best four models. A simple approach to ensemble learning is to take the average of the predictions of the models, where each ensemble member contributes an equal amount to the final predictions. A more sophisticated approach is to take the weighted average of the predictions of the models, in which the contribution of the best model to the final predictions is more than the other models. In this paper, we investigate both ensemble methods on the best four models. Note that the best four models are chosen based on their action classification accuracy. Furthermore, to improve the action classification accuracy, we train our best model on the cropped version of the Stanford 40 dataset, which is obtained by using the bounding box coordinates. These coordinates, which are provided by the dataset creators, show the location of the target person. After training our best model on the cropped version of the Stanford 40 dataset, we apply weighted averaging on this model and the best four models trained on normal version of the Stanford 40 dataset. We expect that the model trained on the cropped version of the Stanford 40 dataset, has a beneficial effect on the ensemble results and thus leads to improving the classification accuracy. Our intuition is that this model has been exclusively trained on the images which contain only the target person, and thus it has a different understanding of actions compared to the models trained on the normal version of the Stanford 40 dataset.

\section{Experimental Results}
\subsection{Implementation details}
We conduct our experiments in the Google Colaboratory environment. Our proposed approach is implemented in Keras~\cite{chollet2015keras} and is evaluated on the Stanford 40 dataset~\cite{yao2011human}, which contains 40 different types of human actions. This dataset has 4000 training images and 5532 test images. We resize all input images to $512 \times 512$ pixels for training and testing. In order to reduce overfitting, four types of data augmentations are randomly adopted: rotation (range of 0-23 degrees), horizontal flipping, width shifting (0 up to 20$\%$), and height shifting (0 up to 20$\%$). To train the proposed model, we set the learning rate to 0.0001 and we use SGD~\cite{bottou2010large} with a momentum coefficient 0.9.

\begin{table}[t]
\centering
\caption{performance of our method with different pre-trained CNNs.}
\label{tab:DiffCNN}
\resizebox{\linewidth}{!}{%
{\renewcommand{\arraystretch}{1.4}
\begin{tabular}{@{}lcc@{}}
\toprule
Our Method with Different Pre-trained CNNs                                      & \begin{tabular}[c]{@{}c@{}}Action Classification\\ Accuracy ($\%$)\end{tabular} & \begin{tabular}[c]{@{}c@{}}Number of \\ Parameters\end{tabular} \\ \midrule
VGG-16~\cite{simonyan2014very}                                  & 72.61                                                                        & 15,118,952                                                      \\
ResNet50~\cite{he2016deep}                                      & 85.39                                                                        & 24,778,408                                                      \\
InceptionV3~\cite{szegedy2016rethinking}                        & 88.81                                                                        & 22,993,480                                                      \\
InceptionResNetV2~\cite{szegedy2017inception}                   & 89.71                                                                        & 55,265,288                                                      \\
DenseNet201~\cite{huang2017densely}                             & 86.08                                                                        & 19,447,144                                                      \\
Xception~\cite{chollet2017xception}                             & 88.83                                                                        & 22,052,176                                                      \\
NASNet-Mobile~\cite{zoph2018learning}                           & 85.67                                                                        & 4,952,508                                                       \\
NASNet-Large~\cite{zoph2018learning}                            & 91.47                                                                        & 87,123,322                                                      \\
NASNet-Large~\cite{zoph2018learning} trained on cropped dataset & 83.92                                                                        & 87,123,322                                  \\ \bottomrule
\end{tabular}%
}
}
\end{table}

\begin{table}[!hb]
\centering
\caption{The Ensemble Learning results.}
\label{tab:Ens}
\resizebox{\linewidth}{!}{%
{\renewcommand{\arraystretch}{1.4}
\begin{tabular}{@{}lc@{}}
\toprule
Ensemble Method                                & \begin{tabular}[c]{@{}c@{}}Action Classification\\ Accuracy ($\%$) \end{tabular} \\ \midrule
Averaging on the Best Four Models                  & 92.67                                                                    \\
Weighted Averaging on the Best Four Models         & 93                                                                       \\
Weighted Averaging on the Best Four Models+Cropped & 93.17                                                                    \\ \bottomrule
\end{tabular}%
}
}
\end{table}

\begin{table}[!hb]
\centering
\caption{Comparison with other methods.}
\label{tab:Com}
\resizebox{\linewidth}{!}{%
{\renewcommand{\arraystretch}{1.4}
\begin{tabular}{@{}lc@{}}
\toprule
Method                                                    & \begin{tabular}[c]{@{}c@{}}Action Classification\\ Accuracy ($\%$)\end{tabular} \\ \midrule
Action-Specic Detectors~\cite{khan2015recognizing}        & 75.5                                                                     \\
Action Mask~\cite{zhang2016action}                        & 82.6                                                                     \\
VLAD spatial pyramids~\cite{yan2017action}                & 88.5                                                                     \\
Multi-branch Attention Networks~\cite{yan2017multibranch} & 90.7                                                                     \\
Part Action Network~\cite{zhao2017single}                 & 91.2                                                                     \\
Ours(Our Model with NASNet-Large)                         & 91.47                                                                    \\
Ours(Averaging on the Best Four Models)                       & 92.67                                                                    \\
Ours(Weighted Averaging on the Best Four Models)              & 93                                                                       \\
Ours(Weighted Averaging on the Best Four Models+Cropped)      & 93.17                                                                    \\ \bottomrule
\end{tabular}%
}
}
\end{table}

\subsection{Performance of Our Method with Different Pre-trained CNNs}
We train our model with eight different pre-trained CNNs on Stanford 40 training dataset. The results on Stanford 40 test dataset and the number of parameters of each model are shown in Table~\ref{tab:DiffCNN}. We can see that our model with NASNet-Large have the best performance in terms of the classification accuracy compared to our model with other pre-trained CNNs. Furthermore, as previously explained, we select the best four models to use them in the Ensemble Learning. In Table~\ref{tab:DiffCNN}, we can observe that our model with NASNet-Large, InceptionResNetV2, Xception, and InceptionV3 are the best four models.

\subsection{Ensemble Learning Results}
As previously described, we apply two types of Ensemble Learning to the top four models, namely Averaging and Weighted Averaging. In Averaging, the ensemble members have equal weights, whereas in Weighted Averaging the best model has the weight equal to 2 and the other three models have the weight equal to 1.  The results of these two types of Ensemble Learning are shown in Table~\ref{tab:Ens}. We can see that the Weighted Averaging leads to better performance, which means giving more weight to the best model has a beneficial effect on the final results. Furthermore, we train our best model, which uses NasNet-Large, on the cropped version of the Stanford 40 dataset and report the test accuracy in Table~\ref{tab:DiffCNN}. Then we apply Weighted Averaging on this model and the best four models trained on the normal version of the Stanford 40 dataset. As it can be seen in Table~\ref{tab:Ens}, the performance improves from $93\%$ to $93.17\%$, which demonstrates the effectiveness of the proposed approach. 

\subsection{Comparison with Previous Methods}
We compare the performance of our approach with previous methods in Table~\ref{tab:Com}. We can see that our method with different settings, performs favorably against the previous methods. The great performance of our method owes to i) using an effective attention mechanism, ii) using NASNet-Large  in the action recognition, and iii) using the Ensemble Learning technique, which combines the predictions of the different models.

\section{Conclusion}
In this paper, we propose using pre-trained CNNs to handle the lack of massive labeled data. Our Feature Extraction part is able to extract more attentive features by benefiting from the Channel Attention Module which is used on top of the CNN. Furthermore, we investigate eight different pre-trained CNNs in our framework. Finally, by combining the predictions of different models, we boost the overall performance of action recognition. Our method achieves 93.17$\%$ accuracy on the Stanford 40 dataset.

{\small
\bibliographystyle{ieee}
\bibliography{paper}
}

\end{document}